\documentclass[letterpaper]{article} 
\usepackage{aaai25}  
\usepackage{times}  
\usepackage{helvet}  
\usepackage{courier}  
\usepackage[hyphens]{url}  
\usepackage{graphicx} 
\usepackage{natbib}  
\usepackage{caption} 
\usepackage{algorithm}
\usepackage{algorithmic}

\usepackage{newfloat}
\usepackage{listings}
\DeclareCaptionStyle{ruled}{labelfont=normalfont,labelsep=colon,strut=off} 
\floatstyle{ruled}
\newfloat{listing}{tb}{lst}{}
\floatname{listing}{Listing}
\usepackage{amsmath}
\usepackage{amssymb}
\usepackage{booktabs}
\usepackage{multirow}
\usepackage{enumitem} \setlist[itemize]{noitemsep}

\usepackage[capitalize, noabbrev]{cleveref}
\crefname{section}{Sec.}{Secs.}
\Crefname{section}{Section}{Sections}
\Crefname{table}{Table}{Tables}
\crefname{table}{Tab.}{Tabs.}
\crefname{figure}{Fig.}{Figs.}

\newcounter{alphaSection}

\title{PLATYPUS: Progressive Local Surface Estimator for Arbitrary-Scale \\ Point Cloud Upsampling}
\author{
    Donghyun Kim\textsuperscript{\rm 1},
    Hyeonkyeong Kwon\textsuperscript{\rm 2},
    Yumin Kim\textsuperscript{\rm 1},
    Seong Jae Hwang\textsuperscript{\rm 1}\thanks{Corresponding author.}
}
\affiliations{
    \textsuperscript{\rm 1}Yonsei University,
    \textsuperscript{\rm 2}Korea University\\

    \{danny0103, yumin, seongjae\}@yonsei.ac.kr, rhaxod9513@korea.ac.kr
}

\begin{document}

\maketitle
\begin{abstract}

3D point clouds are increasingly vital for applications like autonomous driving and robotics, yet the raw data captured by sensors often suffer from noise and sparsity, creating challenges for downstream tasks. Consequently, point cloud upsampling becomes essential for improving density and uniformity, with recent approaches showing promise by projecting randomly generated query points onto the underlying surface of sparse point clouds. However, these methods often result in outliers, non-uniformity, and difficulties in handling regions with high curvature and intricate structures. In this work, we address these challenges by introducing the Progressive Local Surface Estimator (PLSE), which more effectively captures local features in complex regions through a curvature-based sampling technique that selectively targets high-curvature areas. Additionally, we incorporate a curriculum learning strategy that leverages the curvature distribution within the point cloud to naturally assess the sample difficulty, enabling curriculum learning on point cloud data for the first time. The experimental results demonstrate that our approach significantly outperforms existing methods, achieving high-quality, dense point clouds with superior accuracy and detail.

\end{abstract}

\section{Introduction}
\label{sec:introduction}
\begin{figure}[t]
    \centering
    \includegraphics[width=\linewidth]{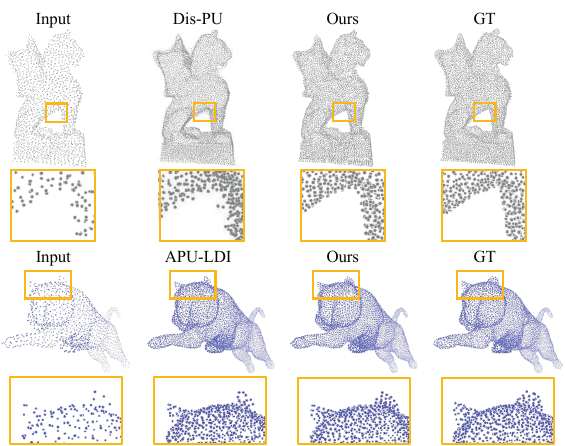}
    \vspace{-15pt}
    \caption{Comparative visualization of 4$\times$ point cloud upsampling results on PU1K. Point cloud upsampling is the task of generating a denser point cloud (i.e., rightmost column) that accurately reflects the underlying geometry of a sparse point cloud (i.e., leftmost column). Our method successfully upsamples intricate areas where existing methods struggle to perform well.}
    \label{fig:previous_problems}

    \vspace{-10pt}
    
\end{figure}

Recently, autonomous driving, robotics, and other technologies that utilize 3D data have attracted significant interest, leading to the growing popularity of 3D point clouds as a representation of 3D data.
However, the raw point clouds captured by sensors such as LiDAR, depth cameras often contain significant noise, and the distribution becomes particularly sparse for points that are farther from the sensor.
Therefore, \textit{point cloud upsampling}, the task of increasing the density points of a sparse 3D point cloud to be of dense (e.g., Fig.~\ref{fig:previous_problems} sparse Input to dense GT) is vital for effectively using raw data in tasks like classification and segmentation, leading to various methods to tackle this challenge.

Starting with early works~\cite{alexa,lipman,huang1} that employed optimization-based methods, the rise of deep learning has led to the proposal of various learning-based methods~\cite{pu_net,pu_gan,pu_gcn}, for training point cloud upsampling networks. Existing learning-based methods have demonstrated excellent performance but they have several drawbacks.
These methods typically split the sparse input point cloud into multiple patches, upsample each patch, and then recombine them. This approach (i.e., \textit{split-and-combine} process), which fails to consider the relationships between patches during upsampling, often results in issues such as holes, outliers, and non-uniformity, especially at the boundaries where the patches are combined.

To address the issues arising from the \textit{split-and-combine} process, recent studies~\cite{grad_pu, apu_ldi} have proposed a pipeline that moves randomly generated points (i.e., query points) onto the surface that the point cloud inherently represents.
In the upsampling process of this pipeline, initial query points are generated around the sparse point cloud (i.e., input point cloud).
Next, these query points are projected onto the underlying surface of the sparse point cloud. 
To accurately determine the underlying surface, it is necessary to know the unsigned distance field of the ground-truth point cloud, but during upsampling, the ground-truth point cloud is not available.
Therefore, in the training phase, the network is trained to predict the unsigned distance from randomly generated query points around the sparse point cloud to its underlying surface. 
This approach enables the network to infer the underlying surface of the dense point cloud using only the sparse point cloud.

\begin{figure}[t]
    \centering
    \includegraphics[width=\linewidth]{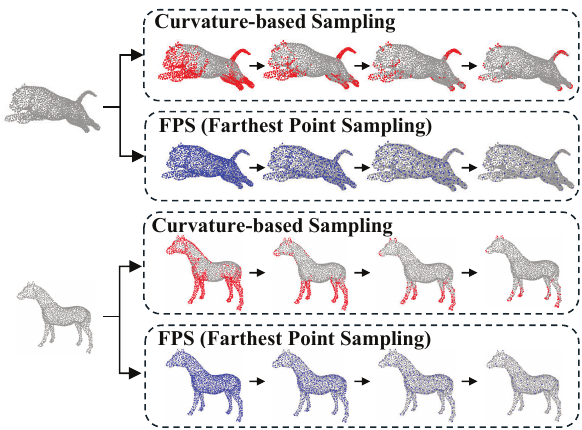}
    \vspace{-15pt}
    \caption{An analysis of the differences between FPS (Farthest Point Sampling) and our newly proposed curvature-based sampling. The comparison shows which points remain as the point cloud is progressively sampled down to fewer points using each sampling method. FPS uniformly samples points across the entire point cloud, whereas curvature-based sampling selectively samples points from regions with intricate structures and high curvature values.}
    \label{fig:fps_vs_cbs}

    \vspace{-10pt}
\end{figure}

While this pipeline has addressed many of the issues in previous methods, there are still some significant problems that remain unresolved.
As seen in \cref{fig:previous_problems}, areas within the point cloud that require high curvature and locally complex structures (e.g., animal paws and ears, object edges) are often poorly upsampled.
To better capture local features in complex regions that existing methods struggle to upsample, our work employs a novel approach using the concept of \textit{curvature value} within the distance-estimating network of this pipeline.
Curvature value represents the degree of curvature in a specific area, calculated by considering the geometric relationships with surrounding points. This metric allows us to quantitatively identify regions where existing methods fail to upsample effectively. 
We then utilize our newly proposed curvature-based sampling technique within the network to explicitly sample these regions with high curvature values (\cref{fig:fps_vs_cbs}).
We refer to this network as \textit{Progressive Local Surface Estimator} (PLSE), which progressively retains regions with high curvature and intricate structures, enabling the network to focus on extracting features from these critical areas.

\begin{figure}[t]
    \centering
    \includegraphics[width=\linewidth]{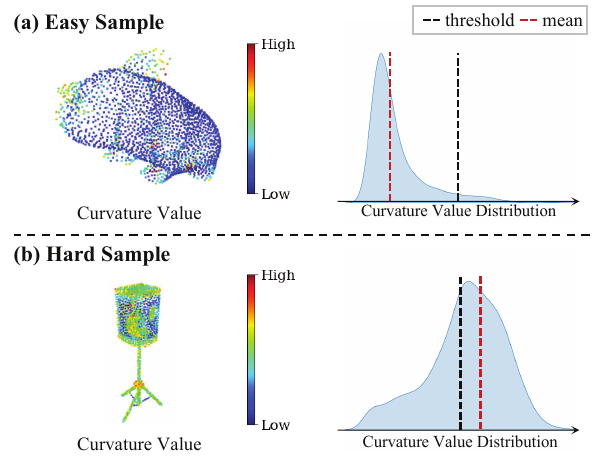}
    \vspace{-15pt}
    \caption{An analysis of the easy samples and hard samples used in our curriculum learning strategy. (a) Point clouds with a higher proportion of points with low curvature values, resulting in a distribution with high skewness, were classified as easy samples. (b) Conversely, point clouds with a higher proportion of points with high curvature values were classified as hard samples.}
    \label{fig:curriculm_learning}

    \vspace{-10pt}
\end{figure}

Using the curvature value, which effectively highlights regions in the point cloud where upsampling is challenging, we further enhance the learning process of PLSE by implementing a new curriculum learning strategy that calculates the difficulty of a point cloud based on the distribution of curvature values.
As shown in \cref{fig:previous_problems}, the network's feature extractor generally struggles more with capturing features in local regions with complex structures and high curvature compared to simpler, flatter areas. 
Based on this observation, if a point cloud has fewer complex structures—indicated by a curvature value distribution skewed toward lower values and a lower mean—we classify it as an easy sample from a learning perspective (\cref{fig:curriculm_learning}). Conversely, if the point cloud's curvature value distribution is skewed toward higher values, indicating more complex structures, it is classified as a hard sample.
Following the curriculum learning strategy, easy samples were used during the early epochs, while hard samples were introduced in the later epochs, helping the network to more effectively learn the unsigned distance field.

\subsubsection{Contributions.} In this work, we introduce \textit{PLATYPUS} (\textbf{P}rogressive \textbf{L}ocal Surface Estimator for \textbf{A}rbi\textbf{T}rar\textbf{Y}-Scale \textbf{P}oint Cloud \textbf{U}p\textbf{S}ampling) for point cloud upsampling.
Specifically, we make the following contributions:
\begin{itemize}
    \item We propose a novel network, Progressive Local Surface Estimator (PLSE), to learn the unsigned distance field from sparse point clouds. PLSE employs a curvature-based sampling method, which allows our network to explicitly focus on extracting features from critical areas.

    \item To improve the learning of the unsigned distance field, we implement a curriculum learning strategy, which classifies training samples into easy and hard based on the skewness of curvature value distribution.

    \item The results from diverse experiments demonstrate that our approach achieves state-of-the-art performance.
\end{itemize}

We provide the code in the supplementary material which will be released upon publication.

\begin{figure*}[t]
    \centering
    \includegraphics[width=\linewidth]{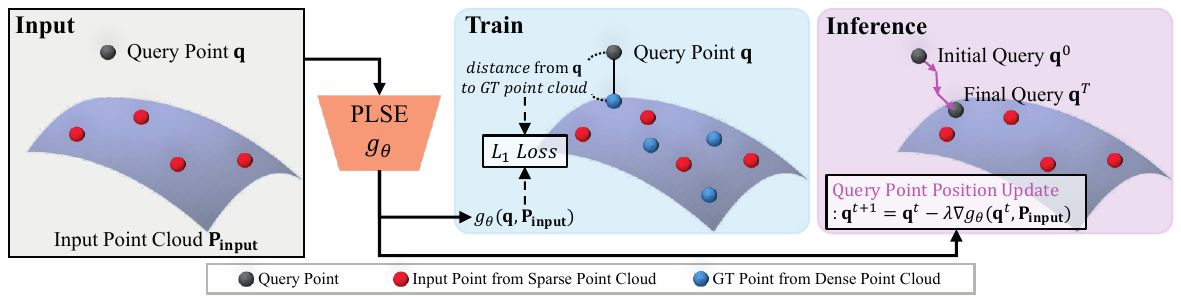}
    \vspace{-15pt}
    \caption{Overall Pipeline of PLATYPUS. \textit{Input}: The input consists of a sparse point cloud $\mathbf{P_{input}}$ and nearby generated query points $\mathbf{q}$. \textit{Train}: During the training process, our Progressive Local Surface Estimator (PLSE) $g_\theta$ is trained to predict the distance from the query point to the underlying surface of the sparse point cloud. The loss compares the distance from the query point $\mathbf{q}$ to the surface of the sparse point cloud (assumed to be identical to the ground truth point cloud). \textit{Inference}: the initial query point $\mathbf{q}^0$ is progressively updates using the PLSE gradient $\nabla g_\theta$ to become the final query $\mathbf{q}^T$ projected onto the surface of the input point cloud.}
    \label{fig:Overall_pipeline}

    \vspace{-10pt}
\end{figure*}

\section{Related Works}
\label{sec:related_works}

\subsubsection{Optimization-based Point Cloud Upsampling.}
As a pioneering approach to point cloud upsampling, Alexa et al.~\citeyear{alexa} proposed interpolating points at vertices of the Voronoi diagram in the local tangent space. Later, Lipman et al.~\citeyear{lipman} introduced the locally optimal projection (LOP) operator for point resampling and surface reconstruction, which was further improved by Huang et al.~\citeyear{huang1} with weighted LOP for iterative normal estimation. However, these LOP-based methods assume points are sampled from smooth surfaces, reducing upsampling quality near sharp edges and corners. To address this, an edge-aware resampling method was introduced, though it relies heavily on normal information and parameter tuning. Subsequently, Wu et al.~\citeyear{wu} proposed a method using Meso-skeleton guidance for consolidation and completion. Overall, optimization-based methods may fail when prior assumptions are unmet.

\subsubsection{Learning-based Point Cloud Upsampling.}
With the rise of deep learning in point cloud analysis, learning-based methods have made significant breakthroughs in upsampling tasks. PU-Net~\cite{pu_net} was the first learning-based method, introducing multi-scale feature learning per point and expanding point sets via multi-branch MLPs. 
However, it overlooked geometric structures, which led to the development of improved methods like the multi-step patch-based progressive network by MPU~\cite{mpu} and PU-GAN~\cite{pu_gan}, which used generative adversarial networks to handle sparse, non-uniform inputs. 
PUGeo-Net~\cite{pugeo_net} introduced a geometry-centric upsampling network, while Dis-PU~\cite{dis_pu} separated the upsampling task into a dense generator and spatial refiner. 
Building on these advancements, recent methods like NePs~\cite{neps} utilized neural fields for high-resolution surface representation, and Grad-PU~\cite{grad_pu} decomposed upsampling into midpoint interpolation and location refinement. APU-LDI~\cite{apu_ldi} introduced an unsigned distance field guided by a local distance indicator for arbitrary-scale upsampling, while RepKPU~\cite{repkpu} used kernel point deformation and cross-attention mechanisms. However, these methods still struggle with capturing local geometry exquisitely, and no methods yet have explored ways to explicitly locally measure the degree of geometric complexity.

\subsubsection{Implicit Neural Representation.}
Implicit neural representation (INR) achieved great performance in 3D shape representation. Conventionally, implicit neural representation works with neural networks by approximating shape functions, such as signed distance functions (SDF)~\cite{deepsdf} or unsigned distance functions (UDF)~\cite{ndf}. Recently, some research in point cloud upsampling has shown advantages by leveraging implicit neural representation for surface representation~\cite{neps, sapcu, pu_ssas, grad_pu, apu_ldi}.
Yet, these INR-based methods innately require per-sample training of the INR networks which quickly becomes impractical when the task demands a fast inference on unseen point clouds.

\section{Methods}
\label{sec:methods}
In this section, we provide a detailed description of our work, PLATYPUS. First, we explain the progression of our upsampling pipeline. Following that, we introduce our novel network, Progressive Local Surface Estimator (PLSE), and our curriculum learning strategy that leverages the distribution of curvature values.

\subsection{Overall Pipeline: Projection-based Upsampling}
The pipeline used by existing methods often generate outliers because the patches are upsampled independently without considering each other.
To address this issue, we follow a projection-based pipeline~\cite{grad_pu, apu_ldi} that predicts the underlying surface of the point cloud and projects points onto the predicted surface (\cref{fig:Overall_pipeline}).

\subsubsection{Upsampling Process.} 
The upsampling process is as follows: query points are randomly generated around the sparse point cloud that needs upsampling. These generated query points are then projected onto the underlying surface of the sparse point cloud using a distance minimization process, similar to the approach used in Grad-PU~\cite{grad_pu}. The distance minimization process utilizes the unsigned distance field of the ground-truth point cloud to move a point to its ground-truth position.
For distance minimization, we need a distance function \( f(\cdot) \) that outputs the shortest distance from a point \( \mathbf{q} \) to the ground-truth point cloud. The distance minimization process can be described by the following equation:
\begin{equation}
    \small
    \mathbf{q}^{t+1} = \mathbf{q}^t - \lambda \nabla f(\mathbf{q}^t, \mathbf{P}_\text{gt}), \quad t = 0, \dots, T-1.
    \label{eq1}
\end{equation}
Here, \( \mathbf{q}^t \in \mathbb{R}^3 \) represents the position of the query point at iteration \( t \), $\mathbf{P}_\text{gt} \in \mathbb{R}^{N \times 3}$ is the ground-truth point cloud of $N$ points, \( \lambda \) is the step size, and \( \nabla f(\mathbf{q}^t, \mathbf{P}_\text{gt}) \) is the gradient of the distance function at \( \mathbf{q}^t \). 
Following this approach, the initially generated query point \( \mathbf{q}^0 \) undergoes several iterations. As a result, the final projected point \( \mathbf{q}^T \) will be positioned on the underlying surface of the ground-truth point cloud.

\subsubsection{Distance Function Training.} 
However, during inference, the ground-truth point cloud $\mathbf{P}_\text{gt}$ is unavailable, which restricts the use of the distance function \( f(\cdot) \) for the distance minimization process.
To address this issue, we train a network $g_\theta(\cdot)$ to predict the shortest distance from a point \( \mathbf{q} \) to the underlying surface of the ground-truth point cloud using only the point \( \mathbf{q} \) and the input point cloud $\mathbf{P}_\text{input}$ (i.e., sparse point cloud). Therefore, we train the network $g_\theta(\mathbf{q}, \mathbf{P}_\text{input})$ which essentially functions as the ideal distance function \( f(\cdot) \) such that $f(\mathbf{q}, \mathbf{P}_\text{gt}) \approx g_\theta(\mathbf{q}, \mathbf{P}_\text{input})$. This, similar to Eq.~\eqref{eq1}, allows the gradient to be approximated using only $\mathbf{P}_\text{input}$ and iteratively projects the input point as follows:
\begin{equation}
    \small
    \mathbf{q}^{t+1} = \mathbf{q}^{t} - \lambda \nabla g_\theta (\mathbf{q}^{t}, \mathbf{P}_\text{input}), \quad t = 0, \dots, T-1.
\end{equation}
Hence, in this projection-based upsampling framework, the quality of the network $g_\theta(\cdot)$ directly dictates the point cloud upsampling quality. In light of this, our work aims to develop a new network for $g_\theta(\cdot)$ which addresses the aforementioned challenges as we describe next.

\begin{figure}[t]
    \centering
    \includegraphics[width=\linewidth]{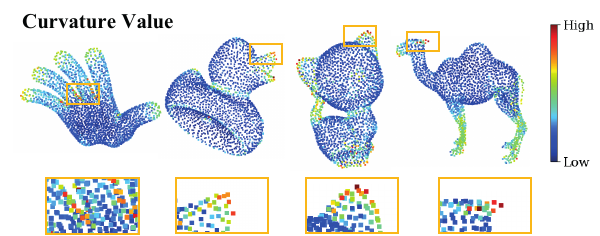}
    \vspace{-15pt}
    \caption{Visualization of each point's curvature value. Points with high curvature values are shown in red, while points with low curvature values are shown in blue.}
    \label{fig:curvature_colormap}

    \vspace{-10pt}
\end{figure}

\begin{figure}[t]
    \centering
    \includegraphics[width=\linewidth]{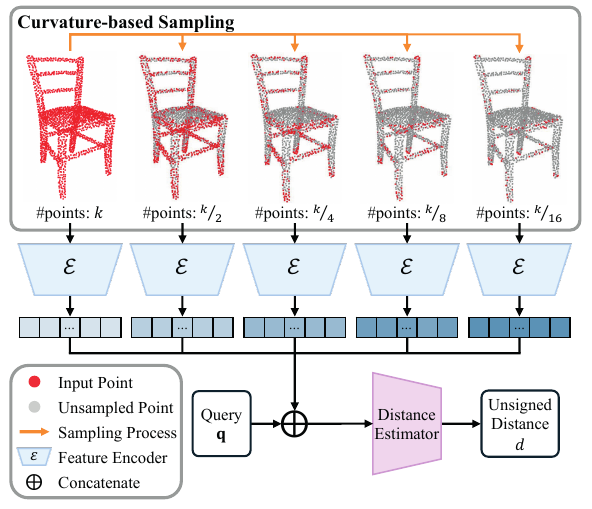}
    \vspace{-15pt}
    \caption{An illustration showing the structure of Progressive Local Surface Estimator (PLSE). The input point cloud with \( k \) points is sampled through curvature-based sampling into point clouds with \( k/2 \), \( k/4 \), \( k/8 \), and \( k/16 \) points. Each of these point clouds passes through a feature encoder to generate features. These features are then concatenated and passed along with the query through the distance estimator, which outputs the unsigned distance \( d \).}
    \label{fig:PLSE_architecture}

    \vspace{-10pt}
\end{figure}

\subsection{Progressive Local Surface Estimator}

We implement a novel network called Progressive Local Surface Estimator (PLSE), which serves the role of $g_\theta(\cdot)$.
This network employs a method that effectively captures local features in regions with high curvature and intricate structures, which existing methods struggle to upsample accurately.

\subsubsection{Curvature Value.} 
We analyze that effectively upsampling challenging areas with intricate structures, such as the edges of objects and the ears of animals, requires focused learning of their local features. 
To achieve this, it is necessary to establish metrics and criteria that quantitatively define these regions.
We find that these structurally complex regions have relatively \textit{high curvature} compared to other areas within the point cloud. By applying the unsigned scalar of umbrella curvature \cite{umbrella_curvature}, we calculate the curvature value \( c_\mathbf{p} \) for each point $\mathbf{p}$ using the following equation:
\begin{equation}
    \small
    c_\mathbf{p} = \frac{1}{K} \sum^K_{i=1} \left| \frac{\mathbf{x}_i}{|\mathbf{x}_i|} \cdot \hat{\mathbf{n}} \right|.
    \label{eq:curvature_value}
\end{equation}
Here, \( \mathbf{x}_i \) is the vector from $\mathbf{p}$ to its neighboring point $\mathbf{p}_i$ (i.e., $\mathbf{x}_i = \mathbf{p}_i-\mathbf{p}$) for each of \( K \) neighboring points. \( \hat{\mathbf{n}} \) is the surface normal vector at $\mathbf{p}$ which is estimated using Open3D~\cite{open3d} due to the absence of the ground-truth surface at $\mathbf{p}$.
Intuitively, based on Eq.~\eqref{eq:curvature_value}, points in regions with complex structures or sharp curvatures will have high curvature values, while points in less complex or flatter areas will have lower curvature values.
As shown in \cref{fig:curvature_colormap}, the calculated curvature values effectively represent the degree of structural complexity, and intricacy of each point in the point cloud. 

\subsubsection{Curvature-based Sampling.} 
We first precompute the curvature value of each point as the criterion for assessing the complexity and detail of the structure. Then, we employ a novel technique called \textit{curvature-based sampling} to ensure the network effectively captures local features in regions with complex structures. As shown in \cref{fig:PLSE_architecture}, curvature-based sampling explicitly selects points with relatively high curvature values from the sparse point cloud, \textit{progressively} sampling point clouds of various sizes over multiple iterations. 
Given a sparse point cloud with \( k \) points, we sample the top \( k/2 \) points with the highest curvature values in the first sampling step. In the subsequent steps, we sample the top \( k/4 \) points, followed by the top \( k/8 \) points, progressively focusing on regions with more complex structures.

\subsubsection{Pipeline of PLSE.}
In our network, called Progressive Local Surface Estimator (PLSE), we extract features \( \mathbf{f}_0 \) through \( \mathbf{f}_4 \) from each of sampled point clouds, where each feature \( \mathbf{f}_i \in \mathbb{R}^d \). The extracted features are then concatenated with the coordinates of the query point \( \mathbf{q} \in \mathbb{R}^3 \), resulting in a final feature vector \( \mathbf{f_\text{final}} \in \mathbb{R}^{3 + d \times 5} \). This final feature vector $\mathbf{f}_\text{final}$ is then passed through the distance estimator module, which outputs the predicted shortest distance from the query points to the underlying surface of the point cloud.
More detailed information about PLSE can be found in the supplementary material.

\subsubsection{Loss Formulation.} 
To ensure that the output of PLSE, the unsigned distance \( g_\theta(\mathbf{q}, \mathbf{P}_\text{input}) \), represents the shortest distance from the query point to the underlying surface of the ground-truth point cloud, PLSE optimizes the L1 loss between its output \( g_\theta(\mathbf{q}, \mathbf{P}_\text{input}) \) and \(f(\mathbf{q}, \mathbf{P}_\text{gt})\) which is distance from the query point \( \mathbf{q} \) to the nearest point in the ground-truth point cloud $\mathbf{P}_\text{gt}$.
This optimization ensures that PLSE accurately predicts the shortest distance from the query point to the underlying surface.

\subsection{Curriculum Learning with Global Curvature Value}
Thus far, we have been using the curvature value to locally characterize each point. However, we realize that the notion of curvature value may easily extend to characterize the global structure complexity of the entire point cloud as a new summary measure, namely, \textit{global curvature value}. This point cloud-level notion of complexity fundamentally allows us to identify \textit{easy samples} (i.e., simple point clouds of points with small $c_\mathbf{p}$) and \textit{hard samples} (i.e., complex point clouds of points with high $c_\mathbf{p}$). Interestingly, this criterion naturally enables \textit{curriculum learning}~\cite{curriculum1, curriculum2} which is a strategy where the network learns samples in order of increasing difficulty, starting with \textit{easy samples} and advancing to \textit{hard samples}. Although curriculum learning is a widely used versatile scheme, we note that applying it to point cloud has not been straightforward due to the absence of sample difficulty measures on point clouds. In our work, our insights on curvature allows us to employ a curriculum learning strategy to aid in the training of our network PLSE.

\subsubsection{Global Curvature Value.} 
To obtain the global curvature value, which is a measure that effectively reflects global structure complexity, we calculate the curvature value for each point in the training sample (i.e., input point cloud) and analyze the distribution of these curvature values. Given a training sample \( \mathbf{P}_\text{input} \) with \( N \) points, where each point \( \mathbf{p}_i \) has a curvature value \( c_{\mathbf{p}_i} \), the global curvature value of \( \mathbf{P}_\text{input} \) is calculated as follows:
\begin{equation}
    \small
    \text{global curvature value} = \frac{1}{N}\sum_{i=1}^N c_{\mathbf{p}_i}.
\end{equation}

We classify the sample's difficulty based on whether the global curvature value is above or below the threshold we set. 
If a training sample has many points with high curvature values and few with low values, it will have a difficulty score above the threshold, classifying it as a hard sample. Conversely, if there are few points with high curvature values and many with low values, the sample will have a difficulty score below the threshold, classifying it as an easy sample.
According to Eq.~\eqref{eq:curvature_value}, each point can have a curvature value between 0 and 1, and we set the threshold at 0.5. Various experiments regarding the threshold setting can be found in the supplementary material.

We use these classified easy samples during the early epochs of the training phase and hard samples during the late epochs to train PLSE. This approach achieves significant performance improvements for PLSE.
\renewcommand{\arraystretch}{1.0}
\setlength{\aboverulesep}{0pt}
\setlength{\belowrulesep}{0pt}

\begin{table}[t]

    \centering
    \resizebox{\linewidth}{!}{

    \begin{tabular}{c c c c c c c}
         \toprule
         \multirow{3}{*}{\centering Method} & \multicolumn{3}{c}{PU-GAN (4$\times$)} & \multicolumn{3}{c}{PU-GAN (16$\times$)}\\

        \cmidrule(lr){2-4} \cmidrule(lr){5-7} 
         
         & CD$\downarrow$ & HD$\downarrow$ & P2F$\downarrow$ & CD$\downarrow$ & HD$\downarrow$ & P2F$\downarrow$ \\

         & $10^{-3}$ & $10^{-3}$ & $10^{-3}$ & $10^{-3}$ & $10^{-3}$ & $10^{-3}$ \\
    
        \midrule

        PU-Net & 0.401 & 4.927 & 4.231 & 0.323 & 5.978 & 5.445 \\ 
        MPU & 0.327 & 4.859 & 3.070 & 0.194 & 6.104 & 3.375\\
        PU-GAN & 0.281 & 4.603 & 3.176 & 0.172 & 5.237 & 3.217 \\ 
        Dis-PU & 0.265 & 3.125 & 2.369 & 0.15 & 3.956 & 2.512 \\
        PU-GCN & 0.268 & 3.201 & 2.489 & 0.161 & 4.283 & 2.632\\
        NePs & 0.385 & 5.615 & 1.642 & 0.147 & 8.851 & 1.925 \\
        Grad-PU & 0.245 & 2.369 & 1.893 & 0.108 & 2.352 & 2.217 \\
        APU-LDI & 0.232 & 1.675 & \textbf{1.338} & 0.092 & 1.504 & \textbf{1.544} \\
        PU-VoxelNet & 0.233 & 1.751 & 2.137 & 0.091 & 1.726 & 2.301 \\
        RepKPU & 0.248 & 2.880 & 1.906 & 0.107 & 3.345 & 2.068  \\

        \midrule

        PLATYPUS & \textbf{0.229} & \textbf{1.426} & 1.908 & \textbf{0.088} & \textbf{1.429} & 2.142 \\

        \bottomrule
    
    \end{tabular}
    }
    \vspace{-3pt}
    \caption{Quantitative comparisons against other methods on the PU-GAN dataset.}
    \label{tab:main_table}

    \vspace{-10pt}
\end{table}
\begin{figure*}[t]
    \centering
    \includegraphics[width=\linewidth]{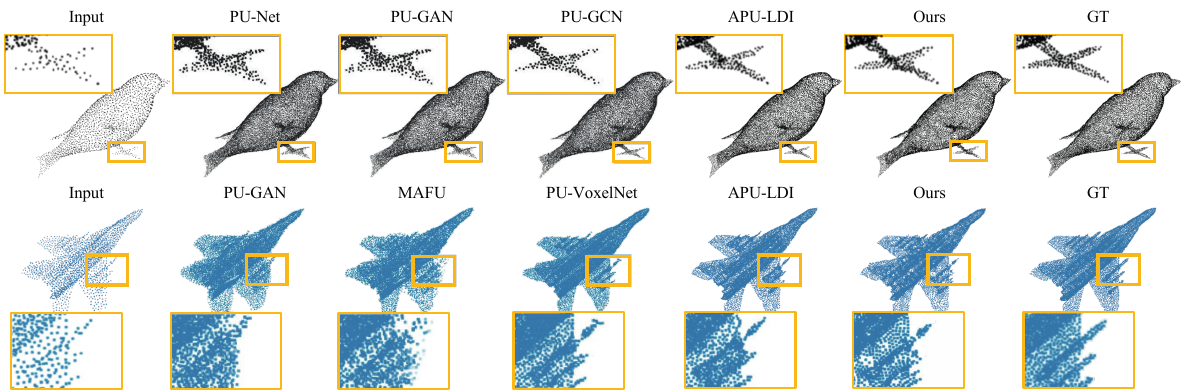}
    \vspace{-15pt}
    \caption{Visualization of upsampling results using synthetic datasets. The top row shows the results on the PU-GAN dataset, while the bottom row presents the results on the PU1K dataset.}
    \label{fig:pugan}
    \vspace{-10pt}
\end{figure*}
\begin{figure}[t]
    \centering
    \includegraphics[width=\linewidth]{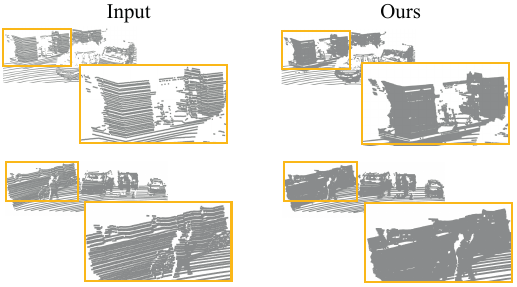}
    \vspace{-15pt}
    \caption{Visualization of upsampling results using KITTI dataset. Objects such as buildings, vehicles, and humans, which are sparsely scanned by the LiDAR sensor, are densely upsampled.}
    \label{fig:KITTI}
    \vspace{-10pt}
\end{figure}

\section{Experiments}
\label{sec:experiments}

\subsection{Experimental Setup}
\subsubsection{Datasets.}
We conducted experiments using two synthetic datasets, PU-GAN~\cite{pu_gan} and PU1K~\cite{pu_gcn}, following the official data split for training and testing. PU-GAN dataset consists of 147 objects, with 120 objects split into 24,000 patches for training, and the remaining 27 objects used for testing. PU1K dataset, which includes more data and a wider variety of objects, consists of 1,147 objects. Of these, 1,020 objects are divided into 69,000 patches for training, while the remaining 127 objects are used for testing. Additionally, to evaluate performance on real-scan datasets, we also conducted experiments using the KITTI dataset~\cite{kitti} and the ScanObjectNN dataset~\cite{scanobjectnn}.

\subsubsection{Implementation Details.}
For our experiments, we used an NVIDIA RTX A6000 GPU. The training process followed a curriculum learning strategy, with 50 epochs for easy samples and 50 epochs for hard samples, totaling 100 epochs, with a batch size of 256. We employed the Adam optimizer during training, with an initial learning rate of 0.001. Additionally, we applied random rotation to augment the training samples. See more details in the supplementary material.

\subsubsection{Metrics and Baselines.}
To evaluate point cloud upsampling performance, we use Chamfer distance (CD), Hausdorff distance (HD), and point-to-surface distance (P2F). The units of CD, HD, and P2F are all $10^{-3}$. We make comparison with various traditional and recent state-of-the-art point cloud upsampling methods, including PU-Net~\cite{pu_net}, MPU~\cite{mpu}, PU-GAN~\cite{pu_gan}, Dis-PU~\cite{dis_pu}, PU-GCN~\cite{pu_gcn}, NePs~\cite{neps}, Grad-PU~\cite{grad_pu}, APU-LDI~\cite{apu_ldi}, PU-VoxelNet~\cite{pu_voxelnet}, and RepKPU~\cite{repkpu}.

\begin{figure}[t]
    \centering
    \includegraphics[width=\linewidth]{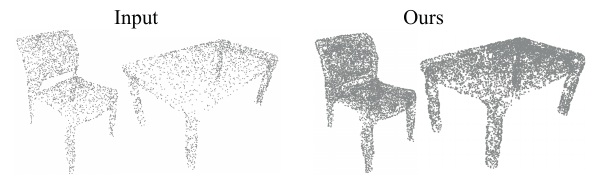}
    \vspace{-15pt}
    \caption{Visualization of upsampling results using ScanObjectNN dataset. Both the real-scanned chair and desk are densely upsampled.}
    \label{fig:scanobjectnn}
    \vspace{-10pt}
\end{figure}

\renewcommand{\arraystretch}{1.0}
\setlength{\aboverulesep}{0pt}
\setlength{\belowrulesep}{0pt}

\begin{table*}[t]
 
    \centering
    

        \begin{tabular}{c c c c c c c c c c}
         \toprule
         \multirow{3}{*}{\centering Method} & \multicolumn{3}{c}{5$\times$} & \multicolumn{3}{c}{7$\times$} & \multicolumn{3}{c}{13$\times$} \\

        \cmidrule(lr){2-4} \cmidrule(lr){5-7} \cmidrule(lr){8-10}

         & CD$\downarrow$ & HD$\downarrow$ & P2F$\downarrow$ & CD$\downarrow$ & HD$\downarrow$ & P2F$\downarrow$ & CD$\downarrow$ & HD$\downarrow$ & P2F$\downarrow$ \\

        & $10^{-3}$ & $10^{-3}$ & $10^{-3}$ & $10^{-3}$ & $10^{-3}$ & $10^{-3}$ & $10^{-3}$ & $10^{-3}$ & $10^{-3}$ \\
    
        \midrule
        MAFU~\cite{mafu}&0.252&2.308&1.949 &0.207&2.593&2.013 &0.177&3.212&2.014 \\
        Grad-PU~\cite{grad_pu}&0.244&2.447&2.459 &0.223&2.654&3.325 &0.242&4.144&5.703 \\
        APU-LDI~\cite{apu_ldi}&0.193&1.706&\textbf{1.445} &0.148&1.534&\textbf{1.468} &\textbf{0.134}&1.878&1.524 \\
        \midrule

        PLATYPUS & \textbf{0.191} & \textbf{1.327} & 1.845 & \textbf{0.137} & \textbf{1.433} & 1.866 & 0.154 & \textbf{1.494} & \textbf{1.379} \\

        \bottomrule
        
        \end{tabular}
    \vspace{-3pt}
    \caption{Quantitative comparisons of arbitrary-scale upsampling performance. The results of upsampling by factors of 5$\times$, 7$\times$, and 13$\times$ were compared with those of other methods.}
    \label{tab:arbitrary_scale_table}

    \vspace{-10pt}
\end{table*}

\subsection{Results on Synthetic Datasets}

\subsubsection{PU-GAN Dataset.}
We conducted experiments on the PU-GAN dataset with upsampling rates of 4 and 16. \Cref{tab:main_table} shows that our PLATYPUS outperforms other methods across various metrics at both upsampling rates. Additionally, the top row of \cref{fig:pugan} presents the upsampling results of PLATYPUS compared to other methods on the PU-GAN dataset. Notably, in the bird's foot area, PLATYPUS produces fewer outliers and preserves the silhouette of the foot compared to other methods.

\subsubsection{PU1K Dataset.}
The bottom row of \cref{fig:pugan} shows our results on the PU1K dataset. When examining the upsampling results for the intricate structures on the wings of the jet, it is evident that PLATYPUS preserves the structure and arrangement of the missiles compared to other methods. Additionally, the quantitative results for PU1K dataset can be found in the supplementary material.

\renewcommand{\arraystretch}{1.0}
\setlength{\aboverulesep}{0pt}
\setlength{\belowrulesep}{0pt}

\begin{table}[t]

    \centering
    \resizebox{\linewidth}{!}{

        \begin{tabular}{c c c c c c c}
             \toprule
             \multirow{3}{*}{\centering Method} & \multicolumn{3}{c}{$\tau = 0.01$} & \multicolumn{3}{c}{$\tau = 0.02$}\\
    
            \cmidrule(lr){2-4} \cmidrule(lr){5-7} 
             
             & CD$\downarrow$ & HD$\downarrow$ & P2F$\downarrow$ & CD$\downarrow$ & HD$\downarrow$ & P2F$\downarrow$  \\
    
            & $10^{-3}$ & $10^{-3}$ & $10^{-3}$ & $10^{-3}$ & $10^{-3}$ & $10^{-3}$ \\
        
            \midrule

            PU-Net & 0.588&6.182&9.842&1.057&9.954&16.282 \\
            PU-GAN &0.435&7.848&7.300&0.815&9.450&14.246 \\
            Dis-PU &0.430&6.580&6.954&0.776&8.861&13.934 \\
            PU-GCN &0.411&5.001&6.963&0.781&8.926&13.730 \\
            {PC}$^2$-PU &0.369&4.390&5.646&0.733&7.921&12.610 \\
            Grad-PU &0.423&4.307&6.403&0.730&6.993&11.481 \\
            APU-LDI & \textbf{0.339}&3.089&\textbf{5.167}&0.622&6.485&10.984 \\
    
            \midrule
    
            PLATYPUS & 0.368 & \textbf{2.224} & 6.804 & \textbf{0.622} & \textbf{5.504} & \textbf{10.633} \\
            
            \bottomrule
        
        \end{tabular}
}
    \vspace{-3pt}
    \caption{Quantitative comparison against other methods for two Gaussian noise levels. \(\tau\) represents the noise level (standard deviation) of the Gaussian noise.}
    \label{tab:noise_table}

    \vspace{-10pt}
\end{table}

\subsection{Results on Real-Scanned Datasets}
To demonstrate that our model, trained on synthetic datasets, generalizes well to diverse real-world scenarios, we conducted upsampling experiments on the outdoor real-scanned KITTI dataset~\cite{kitti} and the indoor real-scanned ScanObjectNN dataset~\cite{scanobjectnn}.

\subsubsection{KITTI Dataset.}
KITTI dataset is related to traffic scenarios used in autonomous driving and robotics and includes point clouds scanned with LiDAR sensor.  We selected point clouds from two different scenes and conducted upsampling experiments. As shown in \cref{fig:KITTI}, our results demonstrate that the our work, PLATYPUS, is well-suited for handling point clouds in real-world scenarios, proving to be a practical and valuable technology.

\subsubsection{ScanObjectNN Dataset.}
ScanObjectNN dataset consists of point clouds scanned from various types of objects. We selected point clouds of chairs and tables to verify whether our method could effectively upsample everyday objects. As shown in \cref{fig:scanobjectnn}, PLATYPUS performed well on real indoor data, demonstrating its effectiveness in these scenarios as well.

\subsection{Additional Analyses}
\label{sec:analysis}

\subsubsection{Arbitrary-Scale Upsampling.}
Most upsampling methods are constrained to the upsampling rate used during training, such as 4$\times$, meaning they can only upsample by 4 times during inference as well. Even when performing inference with a 16$\times$ upsampling rate, it typically involves applying the 4$\times$ upsampling process twice, thereby not exceeding the rate used during training.
To demonstrate that our method can perform arbitrary-scale upsampling, not just upsampling at a fixed rate (4$\times$), we conducted experiments using a model trained on the PU-GAN dataset with a 4$\times$ upsampling rate and applied this model to upsample at rates of 5$\times$, 7$\times$, and 13$\times$. As shown in \Cref{tab:arbitrary_scale_table}, when compared with other methods capable of arbitrary-scale upsampling, PLATYPUS demonstrates excellent performance.

\subsubsection{Robustness on Additive Noise.}
To evaluate whether our model can robustly upsample the correct structure even when the input point cloud's structure is distorted, we conducted experiments by adding Gaussian noise to the xyz coordinates of the input point cloud from PU-GAN test dataset and then upsampling the perturbed point cloud. \Cref{tab:noise_table} presents the results of upsampling with various methods, including ours, using the same noisy input. For both noise levels (i.e., standard deviations), 0.01 and 0.02, PLATYPUS demonstrated superior upsampling performance compared to other methods, indicating that the PLATYPUS upsampling system is robust to variations in the input.

\subsubsection{Ablation Study.}
We conducted experiments to evaluate the effectiveness of various techniques used in PLATYPUS. For the ablation studies, we used the PU-GAN dataset.
\Cref{tab:ablation_table} shows the performance changes in PLATYPUS when the core techniques, curvature-based sampling and curriculum learning based on the skewness of the curvature distribution, are included or excluded. 
When curvature-based sampling is not used, the sparse point cloud bypasses any sampling process, with the entire original point cloud being fed directly into the feature encoder, followed by the distance estimator, which outputs the unsigned distance \( d \). In the absence of the curriculum learning strategy, the training data is not divided into easy and hard samples; instead, all training data is used in every epoch.
As shown in \Cref{tab:ablation_table}, both curvature-based sampling and curriculum learning significantly improve performance compared to the base setting. Moreover, the best performance is achieved when both techniques are applied together. More detailed performance comparisons based on specific settings of curvature-based sampling and curriculum learning can be found in the supplementary material.

\renewcommand{\arraystretch}{1.0}
\setlength{\aboverulesep}{0pt}
\setlength{\belowrulesep}{0pt}

\begin{table}[t]
 
    \centering

        \begin{tabular}{c c | c c c}
         \toprule
         {\centering Curvature-based} & {\centering Curriculum} & CD$\downarrow$ & HD$\downarrow$ & P2F$\downarrow$  \\

        {\centering Sampling} & {\centering Learning}& $10^{-3}$ & $10^{-3}$ & $10^{-3}$ \\
    
        \midrule

        - & - & 0.258 & 2.720 & 2.013 \\
        $\checkmark$ & - & 0.235 & 1.739 & 1.983 \\
        - & $\checkmark$ & 0.249 & 2.521 & 1.948 \\
        $\checkmark$ & $\checkmark$ & \textbf{0.229} & \textbf{1.426} & \textbf{1.908} \\

        \bottomrule
        
        \end{tabular}
    \vspace{-3pt}
    \caption{An ablation study demonstrates the effectiveness of our proposed curvature-based sampling and curriculum learning strategy.}
    \label{tab:ablation_table}

    \vspace{-10pt}
\end{table}
\section{Conclusion}
\label{sec:conclusion}

In this study, we introduce PLATYPUS, a novel upsampling system that addresses the challenges of outliers and the difficulty of upsampling complex regions in point clouds.
Our novel network, Progressive Local Surface Estimator (PLSE), utilizing a newly proposed curvature-based sampling method, effectively captures local features from intricate areas with high curvature in sparse point clouds.
Additionally, the adoption of a curriculum learning strategy allows the network to progressively learn more complex features, leading to better overall performance. 
While there is room for optimizing memory usage, our future work will focus on improving the efficiency of storing sampled point clouds.
Overall, PLATYPUS significantly advances the state-of-the-art in point cloud upsampling, providing a solid foundation for further research and development in this area.

\bibliography{main}

\newpage
\section{Code and Website of PLATYPUS}
\label{sec_supp:code_and_website_of_platypus}

The code for PLATYPUS can be found in the .zip file submitted as supplementary material. Detailed instructions for training and testing are provided in the README.md file. 
Additionally, a project website has been created to introduce PLATYPUS in a simple and accessible manner, which can be accessed at the following link: 
\url{https://platypus-upsampling.github.io/}
\section{Implementation Details}
\label{sec_supp:implementation_details}

\subsection{Detailed Information about Progressive Local Surface Estimator (PLSE)}

\subsubsection{Detailed Flow.}
In Progressive Local Surface Estimator (PLSE), the input point cloud \(\mathbf{P}_\text{input}\) is first sampled into multiple point clouds \(\mathbf{P}_1\), \(\mathbf{P}_2\), \(\mathbf{P}_3\), \(\mathbf{P}_4\) using curvature-based sampling. These sampled point clouds are then passed through the feature encoder \(\mathcal{E}(\cdot)\), extracting features \(\mathbf{f}_0\) through \(\mathbf{f}_4\) as follows:
\begin{equation}
    \mathbf{f}_0 = \mathcal{E}(\mathbf{P}_\text{input}),    
\end{equation}
\begin{equation}
    \mathbf{f}_i = \mathcal{E}(\mathbf{P}_i), \quad i=1,\dots,4.
\end{equation}
These features are concatenated with the query point \(\mathbf{q}\) to create the final feature vector \(\mathbf{f}_\text{final}\), which is then fed into the distance estimator to predict the final unsigned distance \(d\). This process is expressed as follows:
\begin{equation}
    \mathbf{f}_\text{final} = \text{concat}(\mathbf{q}, \mathbf{f}_0, \mathbf{f}_1, \mathbf{f}_2, \mathbf{f}_3, \mathbf{f}_4)
\end{equation}
\begin{equation}
    \ d = \text{distance estimator}(\mathbf{f}_\text{final})    
\end{equation}

\subsubsection{Architecture Details.}
The structure of the feature encoder and distance estimator within PLSE is based on the feature extractor and distance regressor from Grad-PU~\cite{grad_pu}, although many processes and components have been modified.

In Grad-PU, the input point cloud is passed through the feature extractor to generate global and local features. These features are aggregated using a method called feature interpolation, and then concatenated with the query point before being passed to the distance regressor.

In contrast, our work, PLATYPUS, uses a feature encoder similar in structure to Grad-PU's feature extractor. However, the global and local features generated by this encoder are concatenated without undergoing additional processes like feature interpolation. Then, mean pooling is applied to produce each point cloud's feature \(\mathbf{f}_i\) (\(i = 0,\dots,4\)). These features are then concatenated with the query point and passed to the distance estimator as described above.

\subsection{Other Implementation Details}
During the upsampling stage, the query point \(\mathbf{q}\) is projected onto the underlying surface of the point cloud over several iterations. In implementing this process, we compute the feature for a given input point cloud only once before the iterations begin. For each subsequent iteration of the projection, the gradient \(\nabla g_\theta\) is calculated using the fixed feature and the query point \(\mathbf{q}\), which is updated at each iteration.

\section{Quantitative Results on PU1K Dataset}
\label{sec_supp:results_on_pu1k_dataset}

\begin{table}[t]

    \centering

    \begin{tabular}{c c c c}
         \toprule
         \multirow{3}{*}{\centering Method} & \multicolumn{3}{c}{PU1K (4$\times$)} \\

        \cmidrule(lr){2-4} 
         
         & CD$\downarrow$ & HD$\downarrow$ & P2F$\downarrow$  \\

         & $10^{-3}$ & $10^{-3}$ & $10^{-3}$ \\
    
        \midrule

        PU-Net~\cite{pu_net} & 1.157 & 15.297 & 4.924\\ 
        MPU~\cite{mpu} & 0.861 & 11.799 & 3.181\\
        PU-GAN~\cite{pu_gan} & 0.661 & 9.238 & 2.892 \\ 
        Dis-PU~\cite{dis_pu} & 0.731 & 9.505 & 2.719\\
        PU-GCN~\cite{pu_gcn} & 0.585 & 7.577 & 2.499\\
        Grad-PU~\cite{grad_pu} & 0.404 & 3.732 & 1.474\\
        APU-LDI~\cite{apu_ldi} & 0.371 & 3.197 & 1.111\\
        PU-VoxelNet~\cite{pu_voxelnet} & 0.338 & 2.694 & 1.183\\
        RepKPU~\cite{repkpu} & 0.327 & 2.680 & 0.938 \\

        \midrule

        PLATYPUS & 0.412 & \textbf{2.438} & 1.251 \\

        \bottomrule
    
    \end{tabular}
    \caption{Quantitative comparisons against other methods on PU1K dataset.}
    \label{tab:pu1k_table}
\end{table}

As seen in \Cref{tab:pu1k_table}, PLATYPUS demonstrates outstanding performance on the more complex and challenging PU1K dataset, achieving a CD of 0.412, HD of 2.438, and P2F of 1.251. While metrics like CD and P2F are slightly higher compared to other methods, as seen in Fig. 1 of the main paper, PLATYPUS qualitatively outperforms other methods.
\section{Additional Visual Results}
\label{sec_supp:additional_visual_results}

\begin{figure*}[t]
    \centering
    \includegraphics[width=\linewidth]{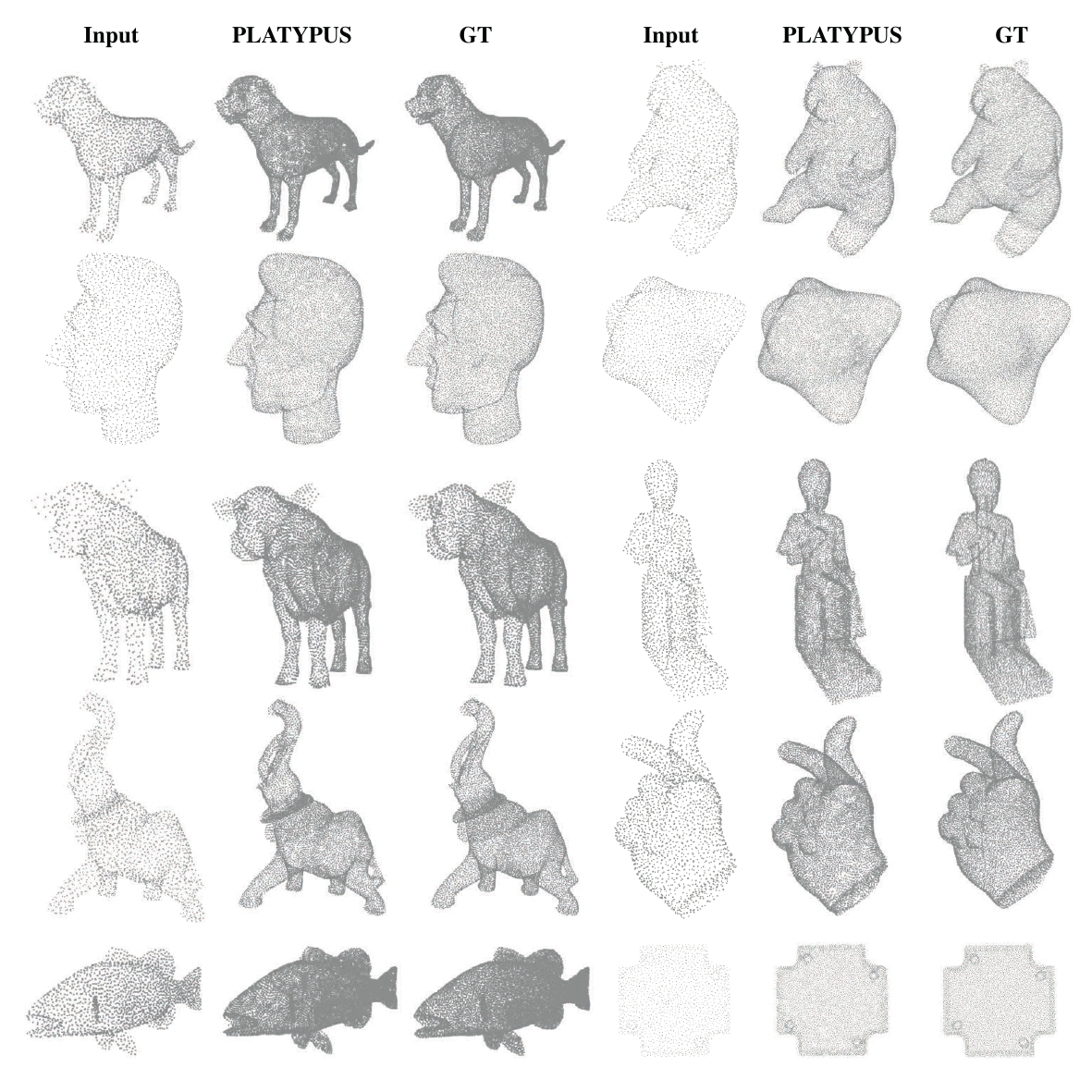}

    \vspace{10pt}
    
    \caption{Upsampling results of PLATYPUS on PU-GAN dataset.}
    \label{fig:pugan_all}

\end{figure*}

\begin{figure*}[t]
    \centering
    \includegraphics[width=\linewidth]{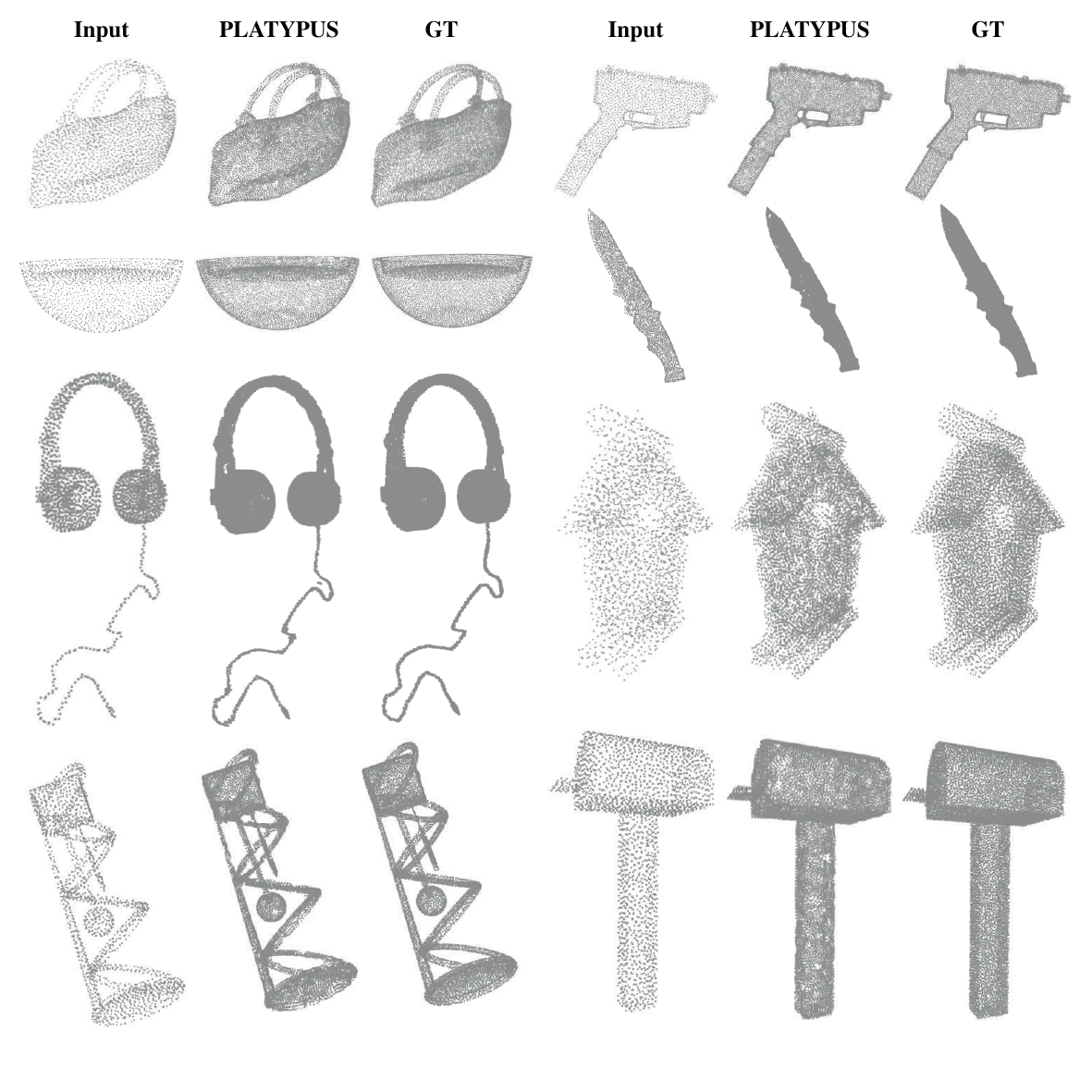}

    \vspace{10pt}
    
    \caption{Upsampling results of PLATYPUS on PU1K dataset.}
    \label{fig:pu1k_all}

\end{figure*}

To accurately and transparently demonstrate PLATYPUS's point cloud upsampling capabilities, we present its visual results on the PU-GAN~\cite{pu_gan} and PU1K~\cite{pu_gcn} datasets. \cref{fig:pugan_all} and \cref{fig:pu1k_all} show the upsampling results of PLATYPUS on the PU-GAN and PU1K datasets, respectively.

\section{Additional Ablation Study}
\label{sec_supp:additional_ablation_study}

\subsection{Steps of Curvature-based Sampling for PLSE}
\begin{table}[t]
 
    \centering

        \begin{tabular}{c c c c}
         \toprule
         \multirow{2}{*}{\centering Sampling Step} & CD$\downarrow$ & HD$\downarrow$ & P2F$\downarrow$  \\

        & $10^{-3}$ & $10^{-3}$ & $10^{-3}$ \\
    
        \midrule
        0 & 0.249 & 2.521 & 1.948 \\
        1 & 0.241 & 2.418 & 1.940 \\
        2 & 0.233 & 2.204 & 1.869 \\
        3 & 0.234 & 1.817 & 1.926 \\
        4 & \textbf{0.229} & \textbf{1.426} & \textbf{1.908} \\
        5 & 0.247 & 1.739 & 1.973 \\
    
        \bottomrule
        
        \end{tabular}

    \caption{Quantitative results based on the number of Curvature-based Sampling steps. The performance varied depending on the number of steps of Curvature-based Sampling used in the Progressive Local Surface Estimator (PLSE) and the number of sampled point clouds utilized. The best performance was achieved with 4 sampling steps.}
    \label{tab:ablation_cbs_table}
\end{table}
In \Cref{tab:ablation_cbs_table}, we evaluated the performance by varying the settings of the core technique, curvature-based sampling. When the curvature-based sampling technique is not applied (i.e., sampling step = 0), the entire unaltered sparse point cloud is fed directly into the feature encoder, and the resulting feature is passed to the distance estimator without any further concatenation, producing the output for the unsigned distance \( d \). Starting from this base setting, we incrementally added sampling steps, evaluating the performance on PU-GAN dataset with up to 5 sampling steps. The experiments showed that using 4 sampling steps yielded the best performance. Based on these results, PLSE is designed to extract features after performing 4 sampling steps.

\subsection{Threshold for Determining Sample Difficulty in Curriculum Learning}
\begin{table}[t]
 
    \centering

        \begin{tabular}{c c c c}
         \toprule
         \multirow{2}{*}{\centering Threshold} & CD$\downarrow$ & HD$\downarrow$ & P2F$\downarrow$  \\

        & $10^{-3}$ & $10^{-3}$ & $10^{-3}$ \\
    
        \midrule
        0.25 & 0.240 & 1.848 & 2.115 \\
        0.5 & \textbf{0.229} & \textbf{1.426} & \textbf{1.908} \\
        0.75 & 0.252 & 1.703 & 2.031 \\

        \bottomrule
        
        \end{tabular}

    \caption{Performance results based on the threshold used for classifying point cloud difficulty in curriculum learning. The point cloud is classified as a hard or easy sample based on whether its global curvature value, which reflects structural complexity, is above or below the threshold. Setting the threshold at 0.5 yielded the best results.}
    \label{tab:ablation_cl_table}
\end{table}
PLATYPUS calculates a value called the global curvature value, which represents the overall structural complexity of the point cloud, to implement its curriculum learning strategy. If this value exceeds the set threshold, the sample is classified as a hard sample; if it is below the threshold, the sample is classified as an easy sample. Since the global curvature value ranges from 0 to 1, we experimented with thresholds set at 0.25, 0.5, and 0.75 on the PU-GAN dataset. \Cref{tab:ablation_cl_table} shows the performance variations of PLATYPUS based on the threshold used in the curriculum learning strategy, with the best performance observed when the threshold is set to 0.5.

\end{document}